\documentclass[conference]{IEEEtran}
\IEEEoverridecommandlockouts
\usepackage{cite}
\usepackage{amsmath,amssymb,amsfonts}
\usepackage{booktabs}
\usepackage{algorithm}
\usepackage{algorithmic}
\usepackage{graphicx}
\usepackage{textcomp}
\usepackage{xcolor}
\usepackage{xspace}
\def\BibTeX{{\rm B\kern-.05em{\sc i\kern-.025em b}\kern-.08em
    T\kern-.1667em\lower.7ex\hbox{E}\kern-.125emX}}

\newcommand{\tann}[1]{\textcolor{black}{#1}}
\newcommand{\sysname}{\textsc{Topas}\xspace}
\newcommand{\jct}{JCT\xspace}
\newcommand{\kv}{KV\xspace}
\newcommand{\paragraphhead}[1]

\title{TOPAS: Workflow-Aware Prefix-State Scheduling for Multi-Agent LLM Serving}
\author{
	\IEEEauthorblockN{
		Hongqiu Ni\IEEEauthorrefmark{1}, 
		Han Tian\IEEEauthorrefmark{1}, 
		Chi Zhang\IEEEauthorrefmark{2}, 
		Guopeng Li\IEEEauthorrefmark{1} 
		and Haisheng Tan\IEEEauthorrefmark{1}} 
	\IEEEauthorblockA{\IEEEauthorrefmark{1}University of Science and Technology of China, China}
	\IEEEauthorblockA{\IEEEauthorrefmark{2}Hefei University of Technology, China}

}

\begin{document}
\maketitle

\begin{abstract}

Prefix caching introduces a fundamental tradeoff in multi-agent large language model (LLM) serving: retaining a long system-prompt key–value (KV) cache for an agent accelerates future calls, yet it reduces the GPU memory available for batching concurrent requests. In multi-stage workflows, existing schedulers tend to prioritize either immediate prefix locality or \tann{overall} workflow progress. However, under a shared KV cache budget, optimizing either objective in isolation can prolong task-level job completion time (JCT)  through downstream delays or frequent prefix replacement. To strike a balance, we here propose \sysname, a Task-Oriented Prefix-Aware Scheduler \tann{that jointly decides which agent prefixes to keep in the cache and which requests to schedule for execution.}  \sysname scores candidate post-decision states \tann{by trading off the expected reduction in each task's longest remaining service path against the near-term benefit of downstream prefix reuse, accounting for the costs of 
prefix movement and preemption. A task‑level aging mechanism is also incorporated to prevent starvation.} We implement \sysname within the SGLang framework and assess its performance on three synthetic DAGs and two MetaGPT software‑development workflows. Compared with the best-performing baseline for each workload and metric, \sysname reduces the mean/p99 JCT by up to
39.8\%/49.4\% on the synthetic
workloads, while lowering mean JCT by 9.8\% on MetaGPT-SOP and mean/p99 JCT by 22.0\%/26.6\% on MetaGPT-TL.

\end{abstract}

\section{Introduction}
\label{sec:intro}

Multi-agent LLM applications tackle complex tasks via interdependent requests generated by role-specialized agents~\cite{autogen,chatdev}. Each request contains an agent-specific static prefix whose KV cache is reusable across requests from the same agent in different tasks. \tann{Serving systems such as SGLang maintain cached KV pairs in a radix tree, where requests sharing the same agent prefix traverse the static path and each appends its own request‑specific dynamic content.} Retaining this cache in GPU memory—referred to as \emph{prefix residency}—eliminates redundant prefill computations, yet it contends with concurrently executing requests for the limited KV budget.  Long system prompts further amplify both the reuse benefit and the residency cost. Consequently, request scheduling influences not only task progress but also the overall serving capacity of the GPU.


Mainstream inference frameworks such as SGLang focus on optimizing batching, prefix
caching, and \kv cache management~\cite{pagedattention,sglang}. Their
schedulers primarily operate at the request level, ranking ready requests
using engine‑local information. Although well suited for per‑request execution optimization, this request‑level view offers only an indirect reflection of the inter‑request dependencies and program‑level progress that the orchestration layer monitors.

Recent efforts aim to bridge this abstraction gap through distinct strategies. Parrot and Autellix incorporate application dataflow or program-level progress into request scheduling \cite{parrot,autellix}, whereas KVFlow leverages workflow context for prefix-cache placement \cite{kvflow}. These systems primarily optimize application progress, cache availability, or
resource placement, while prefix residency remains largely separate from
request admission. This decoupling can leave heterogeneous prefixes co‑residing in the cache—reducing the feasible concurrent batch size—or provoke repeated prefix evictions and reloads as execution shifts across agents. 


\tann{An inherent conflict exists between end-to-end progress against prefix reuse.} Consolidating requests from the same agent allows fewer resident prefixes to support a larger batch, yet it can stall upstream execution while downstream tasks remain pending. Conversely, prioritizing workflow progress tends to trigger frequent agent switching, sacrificing reuse and incurring significant prefix‑movement overhead. Consequently, a scheduler must reconcile prefix locality with holistic progress, rather than optimizing either objective in isolation.

We therefore pose the following question: \textbf{How can a serving system coordinate
prefix residency and request admission throughout workflow execution to minimize
\emph{task-level} job completion time (\jct) under a constrained GPU-memory budget?}
We formalize this problem as \emph{online prefix-state scheduling}. At each
scheduling point, the scheduler jointly decides which agent prefixes to retain and which ready requests to admit, subject to a shared \kv cache budget. This joint decision determines the current batch capacity, immediate task progress, and the set of prefixes most likely to be reused by subsequent workflow steps. 

To solve this problem, we propose \sysname, a Task-Oriented Prefix-Aware Scheduler whose key novelty lies in treating prefix residency as an explicit workflow-level scheduling decision, jointly optimized with request admission, rather than as an indirect byproduct of request ordering and cache eviction.  To our knowledge, \sysname is the first multi-agent LLM scheduler to jointly optimize prefix residency and request admission for task-level \jct under a
shared \kv budget. \tann{\sysname conducts a hierarchical search to construct post-decision GPU states, each specifying a resident-prefix set and a compatible allocation of admitted requests. It selects the optimal state via a JCT-oriented utility that trades off anticipated reduction in each task's longest remaining service path and near-term prefix reuse, accounting for the costs of prefix movement and preemption.}  Our contributions are summarized as:

\begin{itemize}
  \item We formalize the online prefix-state scheduling problem, which jointly selects resident agent prefixes and admitted requests under a shared \kv cache budget. Our diagnostic analysis reveals that heterogeneous prefix co‑residency reduces dynamic batch capacity, \tann{while locality‑first and progress‑first policies fail in complementary leading to bad performance.}
  
\item We introduce \sysname, an online scheduler that jointly determines prefix residency and request admission via a JCT‑oriented utility function. This utility balances expected reductions in tasks' longest remaining service paths and near-term prefix reuse. A task-level aging mechanism is also incorporated to prevent starvation.
  
  \item We implement \sysname atop SGLang and evaluate it on three synthetic DAG
  workloads and two MetaGPT workloads. Compared with the best-performing baseline per workload and metric, \sysname reduces mean/p99 \jct by up to 39.8\%/49.4\% on the synthetic workloads; lowers mean \jct by
  9.8\% on MetaGPT-SOP; and achieves mean and p99 \jct reductions by 22.0\%
  and 26.6\%, respectively, on MetaGPT-TL.
  
\end{itemize}

\section{Motivation: A Progress--Reuse Conflict}
\label{sec:motivation}

\paragraphhead{Why task-level scheduling remains insufficient.}
Task-level policies can use workflow dependencies to decide which tasks or
stages should advance, but their decisions are ultimately executed by a
request-level serving engine. The engine repeatedly admits ready requests into
a continuous batch under a finite \kv budget. Same-agent requests reuse a
resident static-prefix cache, while each request consumes additional \kv
capacity for its dynamic suffix and decode tokens. Admission order and cache
eviction therefore determine both which prefixes remain resident and how many
requests can run together. Task-level priority alone does not control this
prefix--batch interaction; the following diagnostics isolate its two
consequences.

\paragraphhead{Diagnostic 1: heterogeneous prefix co-residency reduces dynamic batch capacity.}
To examine how request order affects batching, we use a two-agent microbenchmark
on the native LLM serving runtime. The workload contains $n$ independent
requests for each of agents $A$ and $B$. The agents have distinct long prefixes,
while all requests use the same dynamic suffix and decode length.
\textsc{Alternating} orders them as $A,B,A,B,\ldots$, keeping both prefixes
resident for most of the run, whereas \textsc{Grouped} serves each agent's
requests together. Both orders contain the same requests and use the same \kv
budget; only their order differs.

\begin{table}[t]
\centering
\small
\begin{tabular}{lccc}
\toprule
Load & \shortstack{Makespan (s)\\(G/A)} &
\shortstack{Mean batch\\(G/A)} & \shortstack{Peak batch\\(G/A)} \\
\midrule
$n=100$ & 1145.5/2086.0 & 9.3/4.7 & 15/8 \\
$n=150$ & 1639.3/3104.4 & 10.0/4.8 & 16/8 \\
$n=200$ & 2143.5/4079.1 & 11.5/5.0 & 16/7 \\
\bottomrule
\end{tabular}
\caption{Under FCFS, heterogeneous prefix co-residency shrinks the running batch, thereby increasing makespan. Slash-separated values are
Grouped/Alternating.}
\label{tab:diag-frag}
\end{table}

Table~\ref{tab:diag-frag} shows that \textsc{grouped} doubles the average
running batch size; \textsc{alternating} takes $1.8$--$1.9\times$ as long to
finish the workload. Serving same-agent requests together lets one resident
prefix support a larger batch. Alternating between agents instead divides \kv
capacity across both long prefixes, leaving less space for dynamic suffixes and
decode tokens.

\begin{figure}[t]
\centering
\includegraphics[width=0.95\linewidth]{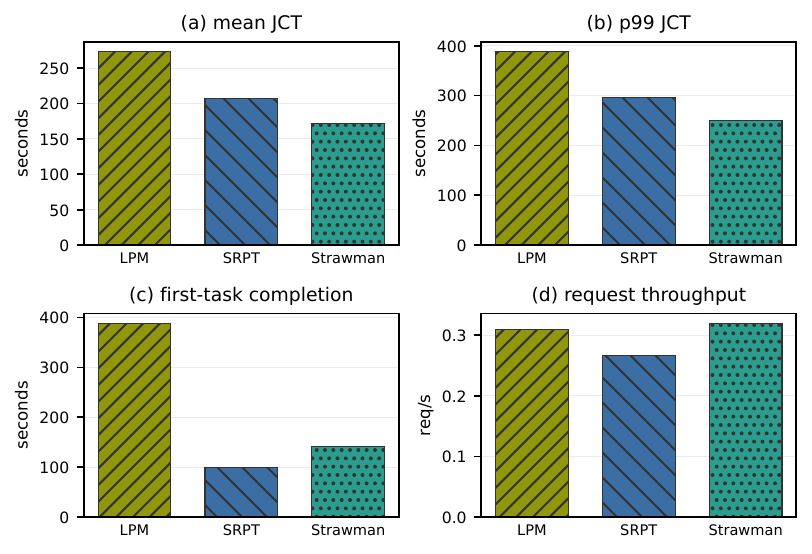}
\caption{Chain-3 at 0.15 task/s under locality-first LPM, progress-first SRPT,
and downstream gating.}
\label{fig:diag-dilemma}
\end{figure}

\paragraphhead{Diagnostic 2: \tann{considering either locality or progress in isolation leads to scheduling failure.}} The batching benefit in Diagnostic 1 suggests a natural response: favor
requests that can reuse a resident prefix. Longest Prefix Match (LPM) is a
representative locality-first policy: it prioritizes ready requests with the longest
immediately reusable prefix. Yet it does not actively choose a target prefix
state. It operates in a reactive, arrival-driven manner:
request order and cache eviction determine prefix residency, which then becomes
the locality signal for the next decision. In an
$A\!\to\!B\!\to\!C$ workflow with continuous arrivals at $A$, this feedback
keeps service near the workflow entrance and delays stages $B$ and $C$. To test
whether active control can break this behavior, we augment LPM with a simple
downstream gate that prioritizes $C$ once its backlog accumulates. Figure~\ref{fig:diag-dilemma} shows that this coarse
intervention achieves lower mean \jct than LPM.

Progress-first scheduling lies at the other extreme. Shortest Remaining
Processing Time (SRPT) prioritizes tasks with shorter remaining processing time
and can move them toward the workflow exit. Yet it has no explicit notion of prefix
residency: repeatedly following the shortest remaining work switches between
agents, causing prefix transfers or recomputation and reducing effective
service capacity. At the same operating point, the downstream-gated policy also
achieves lower mean \jct than SRPT. Thus, neither immediate prefix locality nor
task progress alone is sufficient to minimize task-level \jct.

\begin{figure}[t]
\centering
\includegraphics[width=0.95\linewidth]{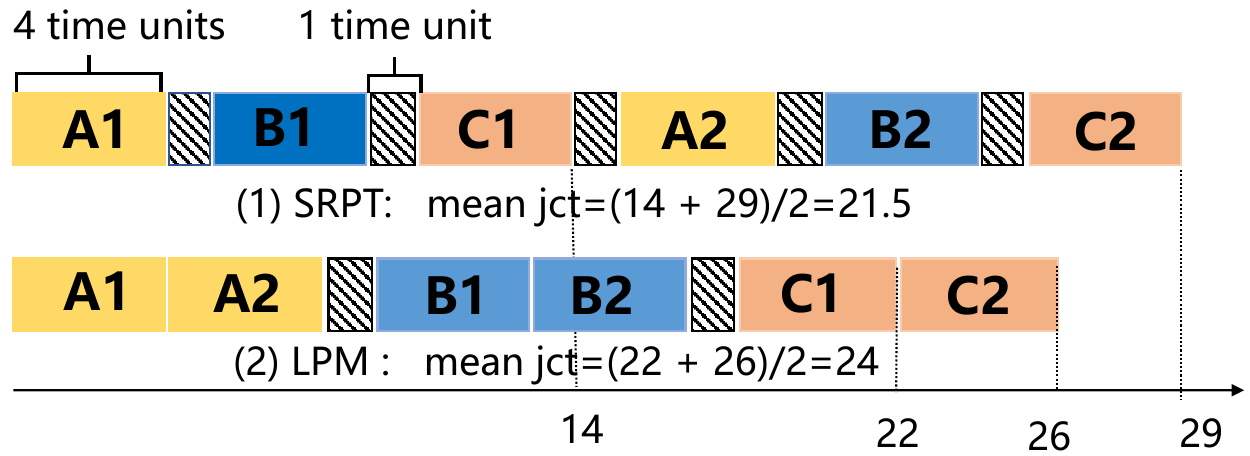}
\caption{A Gantt chart comparing LPM and SRPT on an
$A\!\to\!B\!\to\!C$ workflow. GPU memory can hold the KV cache for only one
Agent prefix at a time; each request takes four time units, and each prefix
setup takes one time unit.}
\label{fig:diag-toy}
\end{figure}

Figure~\ref{fig:diag-toy} exposes the mechanism behind this conflict. SRPT
pushes one task through the chain, but each transition to the next stage changes
the resident prefix and pays another setup cost. LPM instead batches
same-prefix requests to amortize setup, but this locality-first order holds
service at upstream stages and starves downstream progress. The Gantt chart
therefore shows how prioritizing task progress can sacrifice prefix locality,
while preserving locality can delay workflow completion. 

\paragraphhead{Insights.}
An effective scheduler should treat prefix residency as an explicit decision.
It should group requests that share a prefix to preserve batching efficiency,
but move service to another prefix when continued reuse delays task completion.
The goal is to coordinate prefix reuse and task progress within each scheduling
decision.

\section{Problem Formulation}
\label{sec:problem}

\paragraphhead{Workflow and objective.}
All tasks follow the same known DAG $G=(V,E)$. Task $J_i$ arrives at time
$\tau_i$. Each node $v\in V$ is a stage executed by an agent and becomes ready
after all of its predecessors complete. A stage may issue one or more LLM
requests before producing its output. ReAct iterations and tool delays
are not known in advance; they are runtime events that determine when subsequent
requests and stages become ready. Requests produced by agent $a$ share a static
prefix of length $\ell_a$.
Let $C_{i,v}$ be the completion time of stage $v$ in task $J_i$. The task
completion time and task-level \jct are
\begin{equation}
  C(J_i)=\max_{v\in V} C_{i,v},\qquad
  \mathrm{JCT}(J_i)=C(J_i)-\tau_i.
  \label{eq:aaai-jct}
\end{equation}
The serving objective is to minimize average task-level \jct under online
arrivals.

\paragraphhead{Joint prefix--request state.}
At a scheduling point $t$, let $S_t=(R_t,Q_t)$ denote the current GPU state,
where $R_t$ is the set of resident agent prefixes and $Q_t$ is the set of
running requests. A scheduling action selects a post-decision state
$S'=(R',Q')$. Requests in $Q'\setminus Q_t$ are newly admitted, while requests
in $Q_t\setminus Q'$ are preempted. A request can belong to $Q'$ only when its
agent prefix belongs to $R'$. We exclude idle resident prefixes, so
$R'=\{a(r):r\in Q'\}$. The selected joint state must satisfy
\begin{equation}
  \sum_{a\in R'}\ell_a
  +\sum_{r\in Q'} d_r(t)
  \le B_t,
  \label{eq:aaai-budget}
\end{equation}
where $d_r(t)$ accounts for the request's dynamic-suffix \kv, retained output
tokens, and decode reservation, and $B_t$ is the available token-equivalent \kv
capacity. Multiple requests for the same agent pay the static prefix cost once.

\paragraphhead{Online information.}
The scheduler observes ready requests, the current GPU state, task arrival
times, stage progress, and the workflow topology and agent mapping. Future task
arrivals, future ready times, ReAct iterations, and unexpanded tool outputs are
unknown.

\section{\sysname: Workflow-Aware Prefix-State Scheduling}
\label{sec:method}

\subsection{Design Overview}
\label{sec:method-overview}

The two diagnostics in Section~\ref{sec:motivation} point to a limitation
of request-level scheduling. When requests are chosen one at a time, prefix
residency is determined indirectly by request order and the serving runtime's
cache-eviction policy. Different agent prefixes may therefore remain resident
together and leave too little \kv capacity for dynamic suffixes and decoding.
At the same time, service may stay on a reusable prefix even when another
workflow stage matters more to task completion. \sysname instead schedules a
post-decision prefix--request state, directly choosing both the resident
prefixes and the requests served under them.

For each candidate prefix set, \sysname constructs a compatible request
allocation under the joint \kv budget. Its base utility
(Section~\ref{sec:method-utility}) uses the workflow DAG to estimate the
reduction in tasks' remaining LLM-service paths, balancing this progress against
prefix movement and preemption. A one-hop lookahead values prefixes likely to
serve newly released downstream work (Section~\ref{sec:method-reuse}), while
task aging prevents new arrivals from repeatedly overtaking older tasks
(Section~\ref{sec:method-aging}). Section~\ref{sec:method-search} searches these
joint states by enumerating small agent pools and using bounded greedy search
for larger ones. The resulting event-level score uses expected remaining-path
reduction as a heuristic estimate of progress toward task completion. \sysname
greedily commits the highest-scoring generated next state at each scheduling
event and reevaluates the decision as the system state changes.
\subsection{Base Transition Utility}
\label{sec:method-utility}

For a candidate state $S'$, let $A=Q'\setminus Q_t$ and $P=Q_t\setminus Q'$
denote its newly admitted and preempted requests. Candidate feasibility follows
the joint prefix--request capacity constraint in Section~\ref{sec:problem}.
\sysname first evaluates
\begin{equation}
  U_{\mathrm{base}}(S'\mid S_t)
  =J_{\mathrm{admit}}-J_{\mathrm{move}}
   -J_{\mathrm{redo}}-J_{\mathrm{revoke}}.
  \label{eq:base-utility}
\end{equation}
All terms are measured in seconds. $J_{\mathrm{admit}}$ estimates the reduction
in the selected requests' task-level remaining paths; the other terms charge
transition delay and work invalidated by preemption.

\paragraphhead{Task-level admission progress.}
A request's service time is not itself task progress. In a DAG, advancing a
non-bottleneck branch may leave task completion unchanged. \sysname therefore
values an admitted request by the expected reduction in the task's longest
remaining LLM-service path after that request completes. For a chain, the path
is fixed and this value reduces to the expected reduction of the current stage's
residual service. In a fork/join or a general DAG, the bottleneck branch can
change as stages advance, so the value must account for the distributions of
their remaining service.

Under ReAct, let $X_{i,v}\sim F_{a(v),v}$ be the total LLM service of stage $v$
in task $J_i$. Each agent--stage pair has a stage-specific profile of total LLM
service. Once completed executions are available, $F_{a(v),v}$ is their
empirical distribution, including all ReAct rounds but excluding tool waits;
before then, it is a point mass at the profiled total stage-service value. Let
$e_{i,v}(t)$ be the cumulative service recorded in the stage history. A started
but unfinished stage has residual distribution
\begin{equation}
  Z_{i,v}(t)\overset{d}{=}
  X_{i,v}-e_{i,v}(t)
  \;\bigm|\;X_{i,v}>e_{i,v}(t).
  \label{eq:stage-residual}
\end{equation}
Unstarted and completed stages have residuals $X_{i,v}$ and zero, respectively.
Equation~\eqref{eq:stage-residual} conditions repeated ReAct rounds on service
already observed without assuming a fixed number of rounds; tool waits affect
readiness, not this controllable-service estimate.

For request $r$, let $\widehat D_r(t)$ be its expected LLM-service contribution
under the historical service observations for its agent--stage pair. Its
hypothetical completion updates the corresponding residual to
$[Z_{i,v}(t)-\widehat D_r(t)]_+$. For an agent--stage pair without completed
observations, $\widehat D_r(t)$ falls back to the profiled estimate
$T_{\mathrm{prefill}}(|\mathrm{input}_r|)+
T_{\mathrm{decode}}\mathrm{max\_new}_r$ until empirical observations become
available. After completion, the realized service is recorded, and
$F_{a(v),v}$ is updated when the stage finishes.

Let $\mathbf Z_i(\mathcal H,t)$ be the joint residual vector after hypothetical
request completions $\mathcal H$. Conditioned on the observed history, \sysname
models unfinished-stage residuals independently and computes
\begin{equation}
  \widehat L_i(\mathcal H,t)
  =\mathbb E_{\mathbf Z_i(\mathcal H,t)}\!\left[
    \operatorname{LP}\!\left(G,\mathbf Z_i(\mathcal H,t)\right)
  \right],
  \label{eq:expected-remaining-path}
\end{equation}
where $\operatorname{LP}$ is the longest remaining path for a residual
realization, evaluated in reverse topological order. Request $r$ from task $J_i$
then has conditional progress
\begin{equation}
  \Delta_{\mathrm{task}}(r\mid\mathcal H,t)
  =\widehat L_i(\mathcal H,t)
   -\widehat L_i(\mathcal H\cup\{r\},t).
  \label{eq:task-progress}
\end{equation}
The expectation is taken over the empirical conditional residual distributions
of unfinished stages. Placing it outside $\operatorname{LP}$ allows the critical
path to change across residual realizations and stage updates instead of fixing
a path from mean durations.

Within a candidate, requests follow the greedy order constructed in
Section~\ref{sec:method-search}. Let $A_i^{<r}$ contain the earlier hypothetical
request completions from the same task. The base admission term is
\begin{equation}
  J_{\mathrm{admit}}
  =\sum_{r\in A}
   \Delta_{\mathrm{task}}(r\mid A_i^{<r},t).
  \label{eq:j-admit}
\end{equation}
Sequential residual updates make these terms sum to each task's joint expected
remaining-path reduction.

\paragraphhead{Prefix-movement delay.}
\sysname moves prefix \kv caches between GPU and CPU
memory as agents enter and leave the resident set.
Let $R_{\mathrm{in}}=R'\setminus R_t$ and
$R_{\mathrm{out}}=R_t\setminus R'$. Let $\mu$ be the KV bytes per prefix token
and $B_{\mathrm{swap}}$ the swap bandwidth. Transfers within a
transition are serialized, yielding
\begin{equation}
  \begin{aligned}
  T_{\mathrm{move}}(S'\mid S_t)
    &=\frac{\mu}{B_{\mathrm{swap}}}
      \left(\sum_{a\in R_{\mathrm{out}}}\ell_a
      +\sum_{a\in R_{\mathrm{in}}}\ell_a\right),\\
  J_{\mathrm{move}}
    &=T_{\mathrm{move}}(S'\mid S_t)
      |\mathcal T_{\mathrm{in}}|.
  \end{aligned}
  \label{eq:movement-cost}
\end{equation}
Here $\mathcal T_{\mathrm{in}}=\{\mathrm{task}(r):r\in A,
a(r)\in R_{\mathrm{in}}\}$ contains the distinct admitted tasks that wait for an
incoming prefix. Resident-prefix admissions overlap with movement, and each
affected task is charged once.

\paragraphhead{Discarded decode work.}
SGLang reconstructs the generated-token KV of a preempted request through
autoregressive decoding. Let $y_r(t)$ be the number of tokens to reconstruct and
$T_{\mathrm{decode}}$ the profiled decode time per token. The discarded work is
\begin{equation}
  J_{\mathrm{redo}}
  =\sum_{r\in P}T_{\mathrm{decode}}y_r(t).
  \label{eq:j-redo}
\end{equation}

\paragraphhead{Revoked admission credit.}
$J_{\mathrm{redo}}$ charges repeated computation; $J_{\mathrm{revoke}}$ prevents
an unfinished attempt from repeatedly collecting admission value. At admission
time $t_a$, \sysname stores the selected candidate's conditional credit
$\lambda_r=\Delta_{\mathrm{task}}(r\mid A_i^{<r},t_a)$. Preemption revokes it:
\begin{equation}
  J_{\mathrm{revoke}}=\sum_{r\in P}\lambda_r.
  \label{eq:j-revoke}
\end{equation}
The credit remains fixed as sibling branches progress, is cleared on completion,
and is recreated on re-admission. Section~\ref{sec:method-aging} adds aging.

\subsection{Short-Horizon Prefix Reuse}
\label{sec:method-reuse}

Current reuse is already reflected by the capacity constraint: same-agent
requests share one static-prefix cost. To value near-future reuse, \sysname
counts downstream stages about to become ready. Let $n_a(t)$ be the number of
such stages assigned to agent $a$ whose only unfinished predecessor is running;
thus each fork child contributes separately, while a join contributes only when
all other predecessors have finished. A candidate receives
\begin{equation}
  V_{\mathrm{reuse}}
  =\beta\sum_{a\in R'}n_a(t)
   \frac{\mu\ell_a}{B_{\mathrm{swap}}}.
  \label{eq:reuse-value}
\end{equation}
where the reload time $\mu\ell_a/B_{\mathrm{swap}}$ values retaining an
expensive prefix and $\beta$ sets the pressure strength. Lookahead stops after
one workflow transition because farther demand depends on unresolved branches,
ReAct rounds, and tool outcomes. Idle prefixes remain infeasible, so this term
ranks active states rather than prefetching inactive agents.

\subsection{Task-Level Aging}
\label{sec:method-aging}

To keep new arrivals from repeatedly overtaking older tasks with similar
progress, \sysname scales only admission progress by task age:
\begin{equation}
  g_i(t)=1+\rho\log\!\left(
    1+\frac{t-\tau_i}{H_{\mathrm{age}}}
  \right),
  \label{eq:aging-weight}
\end{equation}
where $\tau_i$ is the arrival time of task $J_i$, $H_{\mathrm{age}}$ is the
observed service-time scale used to normalize age, and $\rho$ controls its
strength; the supplementary material gives the update rule. The admission term is
$J_{\mathrm{admit}}^{\mathrm{age}}=\sum_{r\in A}
g_{\mathrm{task}(r)}(t)
\Delta_{\mathrm{task}}(r\mid A_i^{<r},t)$. The stored credit uses the same
admission-time weight, so preemption revokes the value originally granted. The
full transition score is
\begin{equation}
  \mathrm{Score}(S'\mid S_t)
  =J_{\mathrm{admit}}^{\mathrm{age}}
   -J_{\mathrm{move}}-J_{\mathrm{redo}}-J_{\mathrm{revoke}}
   +V_{\mathrm{reuse}}.
  \label{eq:topas-score}
\end{equation}

\subsection{Hierarchical State Search}
\label{sec:method-search}

Direct joint-state search is combinatorial, and task-level utility is not
separable across requests. \sysname therefore separates prefix-set generation
from request allocation. For each set $R$, \textsc{GreedyPack}
retains running requests covered by $R$, derives preemptions, and repeatedly
admits the fitting request with the largest age-weighted conditional progress
per additional request-specific \kv reservation. Same-task marginals are
updated after each choice. It reserves an admission slot and capacity for every
otherwise uncovered prefix in $R$, rejects sets that cannot be covered, and
emits every feasible intermediate allocation. Packing continues through
zero-gain choices to expose complementary fork branches, preferring a represented
task. Let $\Phi(R)$ be the best full score emitted for $R$, or $-\infty$ if none
is feasible.

For the active-agent pool $\mathcal A_t=\{a(r):r\in W_t\cup Q_t\}$, \sysname
enumerates $\mathcal P(\mathcal A_t)$ when $|\mathcal A_t|\le M$. Otherwise it
starts at $R_g=R_t$ and repeatedly takes the best $\Phi$-improving single-agent
addition. At the first stall, it tests all remaining-agent pairs once and, if
one improves $\Phi$, adds the pair and resumes single additions. It also tests
single-agent drop and swap repairs, denoted $\textsc{Repair}(R)$, from $R_t$ and
the final $R_g$, and retains the incumbent. This preserves full prefix-set
enumeration for small pools and limits large-pool search to
$O(|\mathcal A_t|^2)$ prefix-set evaluations.

The cap $K$ applies separately to admissions and preemptions:
$|A_R|\le K$ and $|P_R|\le K$; $K=0$ removes both caps. Natural completions are
committed before search and do not enter $P_R$. From the generated feasible
states $\widehat{\mathcal C}_t$, \sysname selects
\begin{equation}
  S_t^{\mathrm{sel}}
  =\arg\max_{S'\in\widehat{\mathcal C}_t}
   \mathrm{Score}(S'\mid S_t).
  \label{eq:select-generated-state}
\end{equation}
The selection in Eq.~\eqref{eq:select-generated-state} defines an event-level
greedy policy: \sysname executes the highest-scoring feasible candidate and
re-optimizes at the next scheduling event. The score uses expected remaining-path
reduction under the empirical conditional residual distributions as a heuristic
estimate of progress toward task completion, together with the modeled transition
costs and reuse value. \textsc{GreedyPack} enforces feasibility by construction,
and candidate generation bounds the search at every event. With an explicit
event-level score, \sysname evaluates and selects the generated feasible states
directly.
An empty incumbent is removed once a nonempty state is available.

\begin{algorithm}[tb]
\caption{Hierarchical state search}
\label{alg:hierarchical-search}
\textbf{Input}: ready requests $W_t$, incumbent state $(R_t,Q_t)$, budget
$B_t$, and bounds $M,K$\\
\textbf{Output}: selected state $S_t^{\mathrm{sel}}$
\begin{algorithmic}[1]
\STATE $\mathcal A_t\leftarrow\{a(r):r\in W_t\cup Q_t\}$ and
$\widehat{\mathcal C}_t\leftarrow\emptyset$.
\IF{$|\mathcal A_t|\le M$}
  \STATE Add all states from $\textsc{GreedyPack}(\mathcal P(\mathcal A_t))$
  to $\widehat{\mathcal C}_t$.
\ELSE
  \STATE $R_g\leftarrow R_t$ and $p_{\mathrm{pair}}\leftarrow\mathrm{false}$;
  add states from
  $\textsc{GreedyPack}(\{R_t\}\cup\textsc{Repair}(R_t))$.
  \REPEAT
    \STATE Form $\mathcal R_1\leftarrow\{R_g\cup\{a\}:a\in
    \mathcal A_t\setminus R_g\}$ and add
    $\textsc{GreedyPack}(\mathcal R_1)$ states.
    \IF{the best addition improves $\Phi(R_g)$}
      \STATE Update $R_g$ with the best addition.
    \ELSIF{$p_{\mathrm{pair}}=\mathrm{false}$}
      \STATE Set $p_{\mathrm{pair}}\leftarrow\mathrm{true}$; form all
      $R_g\cup\{a,b\}$ once and add their
      $\textsc{GreedyPack}$ states.
      \IF{the best pair improves $\Phi(R_g)$}
        \STATE Update $R_g$ with the best pair.
      \ELSE
        \STATE \textbf{break}.
      \ENDIF
    \ELSE
      \STATE \textbf{break}.
    \ENDIF
  \UNTIL{the search terminates}
  \STATE Add states from $\textsc{GreedyPack}(\textsc{Repair}(R_g))$ and
  add incumbent $S_t$.
\ENDIF
\STATE Remove the empty incumbent if a nonempty candidate exists.
\STATE \textbf{return} $\arg\max_{S'\in\widehat{\mathcal C}_t}
\mathrm{Score}(S'\mid S_t)$ with deterministic ties.
\end{algorithmic}
\end{algorithm}

\section{Experiments}
\label{sec:experiments}

\begin{figure*}[!t]
  \centering
  \includegraphics[width=0.85\textwidth]{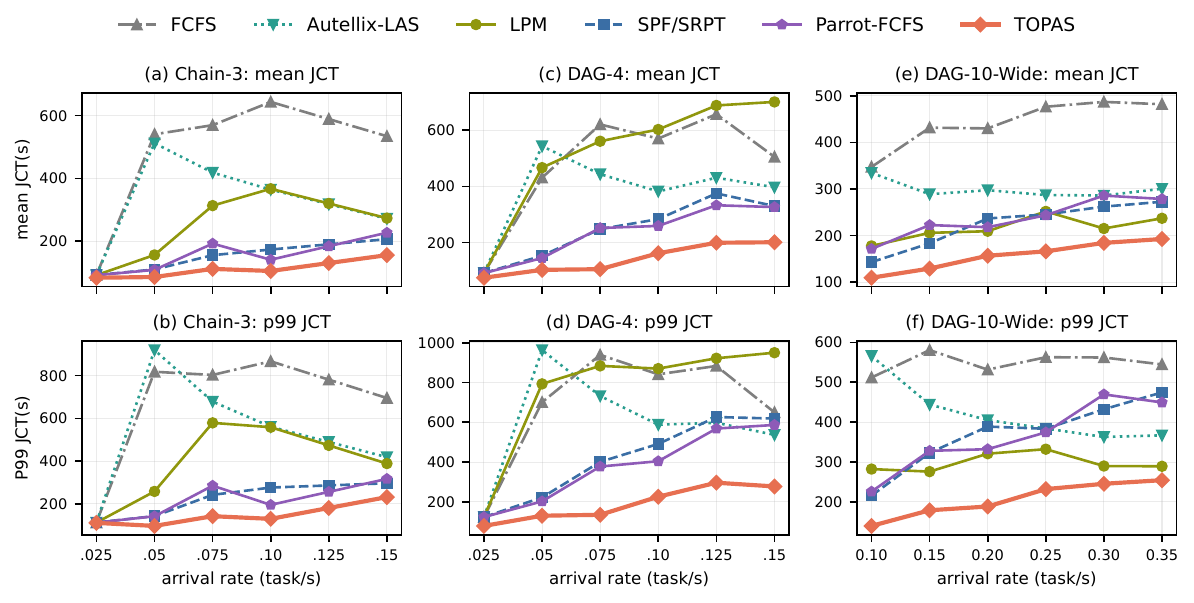}
  \caption{End-to-end task completion on the synthetic DAG workloads.}
  \label{fig:controlled-jct}
\end{figure*}

\subsection{Experimental Setup}
\label{sec:exp-setup}

\paragraphhead{Testbed and Environment.}
We implement \sysname as a scheduling module in SGLang v0.5.3~\cite{sglang}
and evaluate it on a single NVIDIA A100 80GB GPU. The server
allocates the A100 memory remaining after loading the model weights to the
\kv cache; we impose no workload-specific \kv limit. In a separate MetaGPT-SOP
overhead run at 0.15 task/s, \sysname averages 1.9 ms per scheduling decision,
and its cumulative scheduler time accounts for 0.31\% of experiment wall time.

\paragraphhead{Models and Workloads.}
All workloads use Qwen2.5-32B-Instruct. At each operating point, tasks arrive
according to a Poisson process whose rate is the arrival rate shown in the
figures; the sampled arrival trace is fixed and shared across all policies.
We evaluate three synthetic DAG workloads:
\textbf{Chain-3}, \textbf{DAG-4}, and \textbf{DAG-10-Wide}. They cover linear,
fork/join, and wide execution structures with increasing numbers of prefix
namespaces and concurrently ready agents. They use fixed-length user queries
constructed for the evaluation and fixed per-stage generations, isolating
topology and prefix-state pressure from variations in user content.

The two MetaGPT workloads use real prompts sampled
from the MetaGPT SoftwareDev dataset~\cite{metagpt}. The \textbf{MetaGPT-SOP} workload follows the static five-role pipeline
of earlier MetaGPT versions, whereas \textbf{MetaGPT-TL}
unrolls a nine-stage path through the newer star-shaped
organization, alternating a TeamLeader with four specialists to preserve
recurrent reuse of the central prefix.

\paragraphhead{Baselines.}
All policies run on the same SGLang backend. We compare five baselines:
\textbf{(1) FCFS} orders requests by arrival time;
\textbf{(2) Longest Prefix Match (LPM)} prioritizes requests with the
longest reusable prefix; \textbf{(3) Parrot-FCFS}~\cite{parrot} orders ready
requests by task arrival time, using request arrival time as the tie-breaker;
\textbf{(4) Autellix Least-Attained Service (LAS)}~\cite{autellix}
prioritizes the task with the least accumulated LLM service; and
\textbf{(5) Shortest-Path-First (SPF)} prioritizes deeper workflow stages, using
FCFS within a depth. On the equal-cost Chain-3 workload, this ordering is
equivalent to Shortest Remaining Processing Time (SRPT).

\paragraphhead{Metrics.}
We use mean and p99 task \jct as primary metrics. First-task completion time
and request throughput characterize initial progress and serving capacity;
complete results are in the supplementary material. Workload-level comparisons
average each metric over the reported operating points; for each JCT metric,
the comparator is the baseline with the lowest average.

\subsection{End-to-End Task Completion}
\label{sec:exp-overall}

\paragraphhead{Synthetic DAG workloads.}
Figure~\ref{fig:controlled-jct} shows that, relative to the best-performing
baseline for each workload and metric, \sysname reduces mean \jct by 27.5\% on
Chain-3, 39.8\% on DAG-4, and 27.7\% on DAG-10-Wide. The corresponding p99
reductions are 31.7\%, 49.4\%, and 30.8\%.

\begin{figure*}[t]
  \centering
  \includegraphics[width=0.95\textwidth]{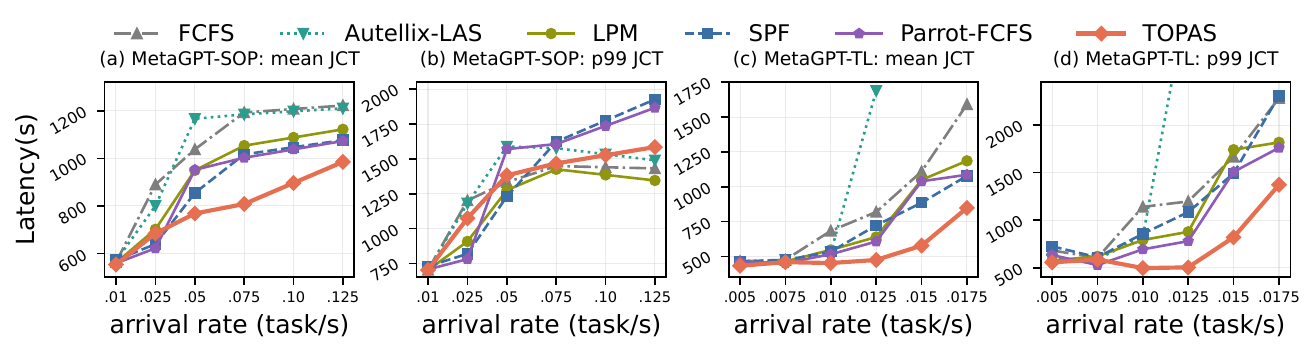}
  \caption{Task completion on the MetaGPT workloads; high-load Autellix-LAS
  values exceed the displayed MetaGPT-TL range.}

  \label{fig:metadata-jct}
\end{figure*}

\paragraphhead{MetaGPT-SOP.}
Figure~\ref{fig:metadata-jct} extends the evaluation to multi-turn traces with
variable call lengths. On MetaGPT-SOP, LPM and Autellix-LAS attain the lowest
p99 \jct and highest request throughput, respectively. Against SPF, the
strongest mean-\jct baseline, \sysname improves all three metrics: it lowers
mean/p99 \jct by 9.8\%/4.5\% and raises request throughput by 6.7\%. 

\paragraphhead{MetaGPT-TL.}
In the TeamLeader trace, a recurrent central prefix competes with heterogeneous
specialist prefixes. At light load, the policies remain close; as contention
grows, \sysname pulls ahead of both locality- and progress-first baselines.
Relative to SPF and Parrot-FCFS, the best-performing baselines for the two
metrics, \sysname lowers mean and p99 \jct by 22.0\% and 26.6\%, respectively.
The difference between SOP and TL follows their prefix dynamics: SOP advances
through five roles in a largely one-pass pipeline, whereas TL repeatedly
returns to a central agent, creating more opportunities for explicit residency
control.


\subsection{Ablations}
\label{sec:exp-ablation}

\begin{figure}[t]
  \centering
  \includegraphics[width=0.95\columnwidth]{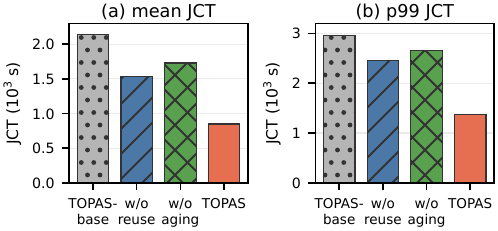}
  \caption{MetaGPT-TL component ablation at 0.0175 task/s.}
  \label{fig:ablation}
\end{figure}

Figure~\ref{fig:ablation} compares the full policy with \sysname-base and the
two single-component ablations on MetaGPT-TL. \sysname-base
retains the base transition utility and joint-state search but omits future
reuse and aging. Relative to \sysname-base, the full policy lowers mean and p99
\jct by 60.5\% and 53.6\%, respectively. Both components contribute: compared
with the two single-component variants, the full policy further lowers mean
\jct by 44.9--51.0\% and p99 \jct by 44.2--48.5\%.

\section{Related Work}
\label{sec:related}

\paragraphhead{LLM serving and prefix reuse.}
LLM servers improve GPU efficiency with efficient attention kernels, continuous
batching, paged \kv-cache allocation, preemption, and chunked
prefill~\cite{flashattention,orca,pagedattention,fastserve,sarathiserve};
disaggregated systems separate prefill and decode~\cite{distserve,splitwise}.
Prefix-aware systems reuse shared computation~\cite{sglang,promptcache,
chunkattention}, route requests toward cached prefixes~\cite{preble,mooncake},
or manage caches across memory tiers~\cite{flexgen,cachedattention,memserve}.
These mechanisms optimize request execution, locality, or data movement without
considering how prefix residency advances a workflow under limited GPU memory.

\paragraphhead{Workflow-aware and agentic serving.}
Agentic applications interleave model calls with tools and coordinate model or
agent modules~\cite{react,toolformer,hugginggpt,chameleon,dspy}. At the serving
layer, Parrot exposes application dataflow; InferCept and Continuum preserve
\kv caches across external waits; and Autellix and Astraea schedule calls by
program progress or lifecycle state~\cite{parrot,infercept,continuum,autellix,
astraea}. AugServe and Teola optimize augmented applications~\cite{augserve,
teola}; KVCOMM and DroidSpeak transfer \kv caches across contexts or
models~\cite{kvcomm,droidspeak}; and multi-agent systems optimize routing, data
access, allocation, batching, and runtime execution~\cite{kairos,helium,
selfresource,halo,flowmesh}.
KVFlow is a workflow-aware KV-cache management system that uses an Agent Step
Graph to guide KV-node eviction and prefetching, temporarily skipping
requests whose caches are still loading~\cite{kvflow}.
\sysname instead
makes the memory contention between resident agent-prefix
caches and running requests part of workflow scheduling, using DAG progress and reuse to reduce task-level \jct.


\section{Conclusion}
\label{sec:conclusion}

Multi-agent LLM workflows couple prefix residency with ready-request admission
under a shared GPU-memory budget. \sysname jointly schedules these decisions,
balancing expected remaining-path reduction and near-term reuse against prefix
movement and preemption. Relative to the best-performing baseline for each
workload and metric, it reduces mean and p99 task \jct on synthetic DAG
workloads by up to 39.8\% and 49.4\%, respectively. It lowers mean \jct on both
MetaGPT workloads, with reductions reaching 22.0\%; on MetaGPT-TL, it also
lowers p99 \jct by 26.6\%.
\bibliographystyle{plain}
\bibliography{aaai2027}

\end{document}